\documentclass[letterpaper, 10 pt, conference]{ieeeconf}  
\IEEEoverridecommandlockouts                              
\usepackage{packages}

\title{\LARGE\bf 

Fast and Robust Localization for Humanoid Soccer Robot \\ via Iterative Landmark Matching 

}
\author{Ruochen Hou$^{1}$, Mingzhang Zhu$^{1}$, Hyunwoo Nam$^{1}$, Gabriel I.~Fernandez$^{1}$, and Dennis W.~Hong$^{1}$% <-this % stops a space
\thanks{
$^{1}$Robotics and Mechanisms Laboratory (RoMeLa), Department of Mechanical and Aerospace Engineering, University of California, Los Angeles, CA 90095, USA.
        {\tt\small \{houruochen, normanzmz, harrynam, gabriel808, dennishong\}@ucla.edu}}
}

\begin{document}
\maketitle
\thispagestyle{empty}
\pagestyle{empty}
%%%%%%%%%%%%%%%%%%%%%%%%%%%%%%%%%%%%%%%%%%%%%%%%%%%%%%%%%%%%%%%%%%%%%%%%%%%%%%%%

\begin{abstract}
% Problem or status quo
Accurate robot localization is essential for effective operation. Monte Carlo Localization (MCL) is commonly used with known maps but is computationally expensive due to landmark matching for each particle. Humanoid robots face additional challenges, including sensor noise from locomotion vibrations and a limited field of view (FOV) due to camera placement. This paper proposes a fast and robust localization method via iterative landmark matching (ILM) for humanoid robots. The iterative matching process improves the accuracy of the landmark association so that it does not need MCL to match landmarks to particles. Pose estimation with the outlier removal process enhances its robustness to measurement noise and faulty detections. Furthermore, an additional filter can be utilized to fuse inertial data from the inertial measurement unit (IMU) and pose data from localization. We compared ILM with Iterative Closest Point (ICP), which shows that ILM method is more robust towards the error in the initial guess and easier to get a correct matching. We also compared ILM with the Augmented Monte Carlo Localization (aMCL), which shows that ILM method is much faster than aMCL and even more accurate. The proposed method's effectiveness is thoroughly evaluated through experiments and validated on the humanoid robot ARTEMIS during RoboCup 2024 adult-sized soccer competition.

\end{abstract}

%%%%%%%%%%%%%%%%%%%%%%%%%%%%%%%%%%%%%%%%%%%%%%%%%%%%%%%%%%%%%%%%%%%%%%%%%%%%%%%%

%% Paper sections to be imported
% \input{Sections/S0_nomenclature}
\section{Introduction}
\label{sec:intro}
% introduce robocup and localization
RoboCup is an international robot soccer competition where accurate localization is crucial for decision-making and path planning \cite{hou2025path}. Real-time localization with limited computational resources requires fast and efficient algorithms, as delays in localization affect path planning and trajectory tracking performance. Therefore, fast and accurate localization is essential for our humanoid platform, ARTEMIS, which can move at speeds up to 1.5 m/s. Since ARTEMIS walks on a 2D plane, the localization problem is inherently simplified to 2D in this paper.

Unlike wheel-based robots, bipedal locomotion and motor vibrations introduce more noise to the sensors. Additionally, to closely mimic human movement, the competition restricts the use of sensors to only the camera mounted on the actuated neck, further complicating the localization challenge. Traditional visual odometry (VO) based methods \cite{oriolo2012vision}, which rely on tracking visual features between consecutive frames, are not suitable due to the presence of moving people and robots on the field, which can significantly affect VO accuracy. Kinematic-based methods are also prone to inaccuracies, especially due to sliding on the grass. Learning based methods \cite{kendall2015posenet}, while effective, require training on specific images from a given location, making them impractical and time-consuming for tournament settings.

\begin{figure}[t]
    \centering
        \includegraphics[width=\linewidth]{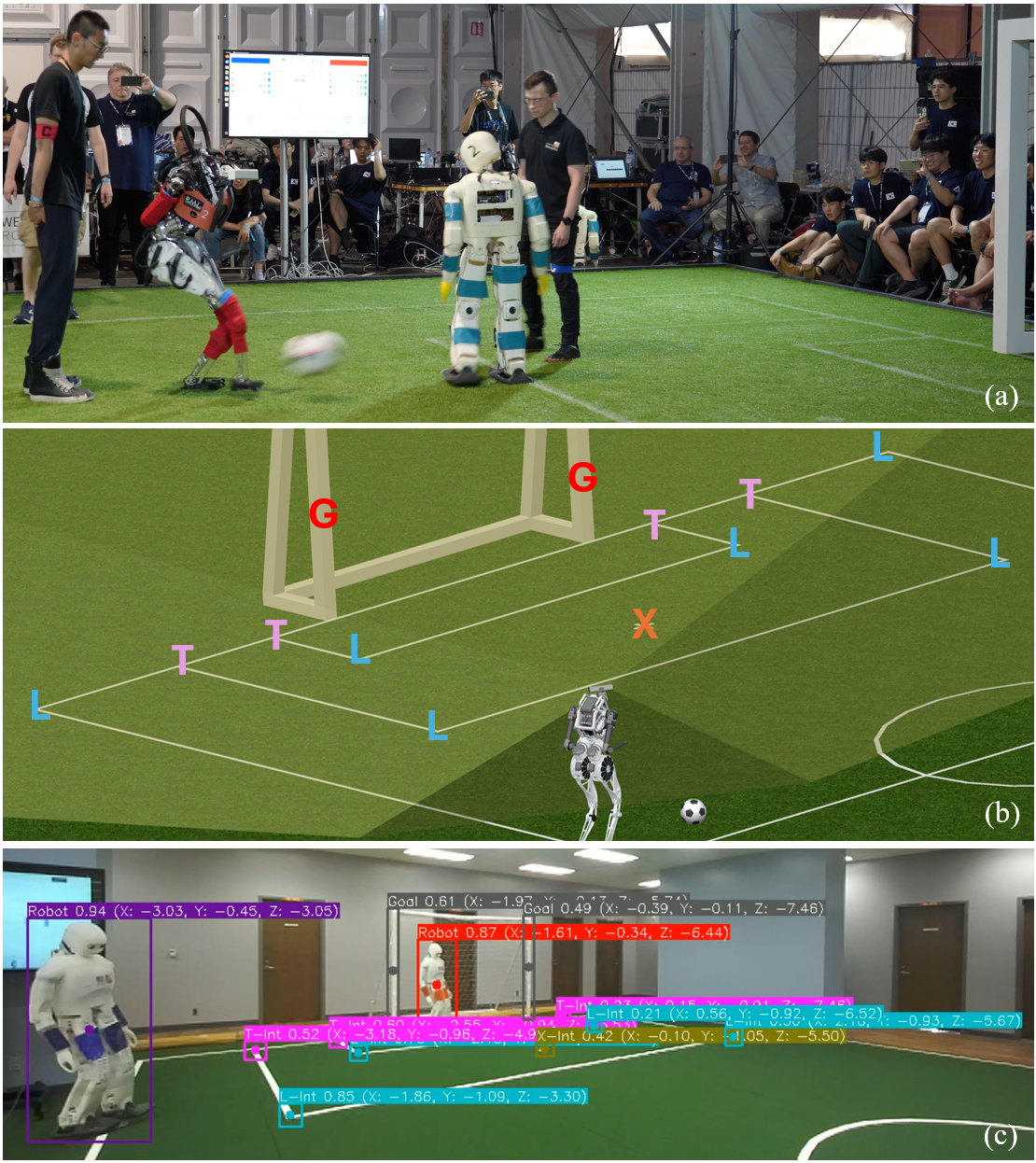}
        % \caption{Artemis shooting A}
        % \caption{}
        % \label{fig:artemis_shooting}

    \caption{(a) Picture from RoboCup 2024's championship match \cite{robocupRomelaSymposium} with our robot, ARTEMIS, shown in red taking a shot on goal from far away. A person dressed in black carries an emergency stop behind ARTEMIS for safety.
    (b) A 3D simulated environment where the yellow polyhedron represents the camera's field of view in 3D. The mapped field features, including goal posts marked with a \textit{red G}, corners denoted with a \textit{green L}, t-intersections with a \textit{pink T}, and crosses with a \textit{yellow X}.
    (c) Illutrates landmarks, goal post and robots detection using the ZED 2i camera.
    % Average pose estimation error in position and orientation considering different levels of noise.
    }
    \label{fig:artemis_shooting}
    % \label{fig:pose_estimating_error}
\end{figure}

% field and landmark intro::
Given that the details of the soccer field are provided ahead of time, we can use the field features as input to estimate the robot's pose. Approaches utilizing field lines have been explored \cite{laue2009efficient, schulz2012utilizing, nagi2014vision, muzio2016monte, ficht2018grown}, as well as methods using pre-defined landmarks such as corners, T-intersections, crosses, and goalposts \cite{kim2023enhancing, dwijayanto2019real}, as shown in \cref{fig:artemis_shooting}. Matching pre-defined landmarks is easier than using lines, as the robot can detect multiple landmarks but typically cannot see the entire line. Furthermore, YOLO-based landmark detection \cite{kim2023enhancing, dwijayanto2019real} is fast and accurate. In this work, we used a YOLOv8 network trained on a custom landmark data set to identify these landmarks.

% \begin{figure}[t!]
%     \centering
%     \includegraphics[width=0.9\linewidth]{Figures/S1_introduction/landmarks.png}
%     \caption{Pre-defined landmarks on the soccer field.}
%     \label{fig:pre_defined_landmarks}
% \end{figure}

% method intro::

A common method for localization on a known map like a soccer field is Monte-Carlo localization (MCL) using the landmarks on the field \cite{kim2023enhancing,thrun2005probabilistic,hong2009robust,hornung2010humanoid,almeida2017vision,nagi2014vision}. The MCL method uses particles to explore the possible states, evaluate them, and assign weights to those particles. However, the MCL method needs to match and evaluate landmarks for each particle, which is computationally heavy and slow, especially for a large number of particles.

\begin{figure}[t]
    \centering
    \includegraphics[width=0.9\linewidth]{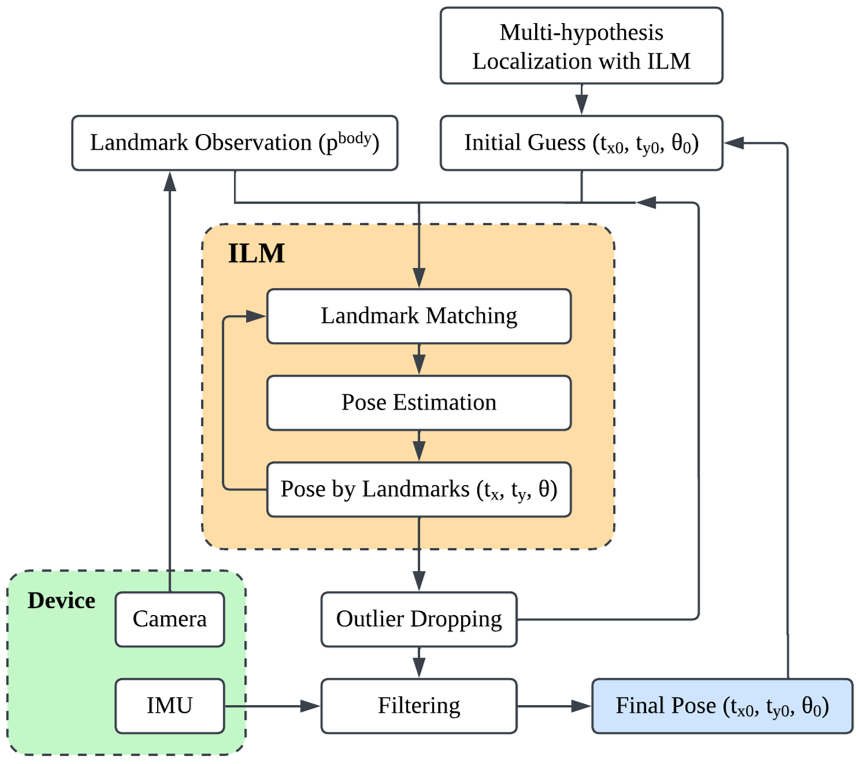}
    \caption{Work flow of our localization method. ILM takes the landmark observation and initial guess for its pose as an input. It then matches and estimates the pose iteratively and drops outliers. The pose from ILM is fused with data from an inertial measurement unit (IMU) using a filter to get the final localized state.}
    \label{fig:work_flow}
\end{figure}

To address this issue, we propose iterative landmark matching (ILM) to improve the matching accuracy. Our method directly calculates the pose from the matched landmarks, eliminating the need for MCL to explore multiple particles. \cref{fig:work_flow} shows the proposed localization framework. Initially, Multi-Hypothesis Localization (MHL) is used to localize the starting pose. ILM takes the landmark observation and the previous pose as an initial guess, iteratively matches and estimates the pose, and removes outliers. The estimated pose is then fused with IMU data through a filter to obtain the localized state of the robot.

% \begin{figure}[htbp]
%     \centering
%     \includegraphics[width=0.9\linewidth]{Figures/S1_introduction/localization_opt_chart.png}
%     \caption{Work flow of our localization method. ILM takes the landmark observation and initial guess for its pose as an input. It then matches and estimates the pose iteratively. The pose from ILM is fused with data from an inertial measurement unit (IMU) to get the final localized state}
%     \label{fig:work_flow}
% \end{figure}

% organization ::
This article is organized as follows: 
In \cref{sec:related_work}, we discuss the existing methods related to our method.  
In \cref{sec:localization_framework}, we present our localization framework.
% including ILM algorithm, multi hypothesis localization, outlier dropping and filtering.  
% In \cref{sec:pose_esti}, we provide two methods for 2D pose estimation and give a detailed comparison for the calculation time and estimation error under noised data.
% In \cref{sec:iterative_landmark_matching}, the ILM algorithm is illustrated and tested. 
% In \cref{sec:data_fusion}, pose data from localization is fused with inertial data from IMU.
The experimental results are presented and discussed in \cref{sec:exp_result}, where we compare ILM with ICP and aMCL, demonstrating its strong performance. All computations were performed on a Dell G16 laptop with an Intel Core i7-13650HX processor. Simulations used a 110-degree field of view (FOV), matching the ZED 2i camera used in the competition.
 
\section{Related Work}
\label{sec:related_work}
% \subsection{Other Methods}
\subsection{Iterative Closest Point}
The Iterative Closest Point (ICP) algorithm, which was introduced by Besl and McKay \cite{besl1992method}, is a widely used method for point cloud registration. It iteratively refines the transformation by establishing correspondences based on the closest points and minimizing the alignment error. The classical point-to-point ICP uses nearest neighbor matching. There are also other variations, such as point-to-plane ICP \cite{chen1992object}, which improves accuracy by incorporating surface normals; Generalized ICP (GICP) \cite{segal2009generalized}, which improves registration by using covariance-based metrics; and Deep Closest Point (DCP) \cite{wang2019deep} and PointNetLK \cite{aoki2019pointnetlk}, which are deep learning-based alternatives.

The ILM method can be considered a modification of ICP. One key difference is that ILM finds one-to-one matching, whereas ICP can result in multiple source points being matched to the same target point. One intuitive reason why ILM performs better than ICP in our case is that the landmarks are identified by YOLOv8. By design, YOLO \cite{redmon2016you} (including YOLOv8 \cite{terven2023comprehensive}) does not assign two bounding boxes to the same object. Non-Maximum Suppression (NMS) is applied to remove redundant overlapping boxes, keeping only the one with the highest confidence score. Therefore, using the nearest point strategy to establish correspondences could incorrectly assign multiple landmarks to the same point on the map, which is typically inaccurate. This makes ICP more prone to getting trapped in local minima.

\subsection{Data Association}
\label{sec:Data_Association}
% problem definition:
Our approach begins by matching the landmarks observed by the stereo vision system to their corresponding landmarks in the known a priori map of the soccer field. The vision system detects landmarks such as corners, T-intersections, crosses, and goal posts relative to the robot. Matching these landmarks to their positions in the map can be formulated as a typical Linear Assignment Problem (LAP), defined as follows:

\begin{itemize}
    \item A cost matrix \( C \in \mathbb{R}^{n \times n} \), where \( C_{ij} \) represents the distance of assigning observed landmark \( i \) to target landmark \( j \) in the map. If \( C \) is not square, it can be made square by adding zero entries.
    \item A binary assignment matrix \( Y \in \{0,1\}^{n \times n} \), where \(Y_{ij} = 1\) if observation \( i \) is assigned to target \( j \), and  \(Y_{ij} = 0\) otherwise. 

\end{itemize}

The goal is to minimize the total assignment cost:

\[
\min_{Y} \sum_{i=1}^{n} \sum_{j=1}^{n} C_{ij} Y_{ij}
\]

Subject to:
\[
\sum_{j=1}^{n} Y_{ij} = 1,
\sum_{i=1}^{n} Y_{ij} = 1,
Y_{ij} \in \{0,1\},\forall i,j \in \{1,\dots,n\}
\]

% \[
% \sum_{j=1}^{n} X_{ij} = 1, \quad \forall i \in \{1, \dots, n\}
% \]
% \[
% \sum_{i=1}^{n} X_{ij} = 1, \quad \forall j \in \{1, \dots, n\}
% \]
% \[
% X_{ij} \in \{0,1\}, \quad \forall i,j \in \{1, \dots, n\}
% \]
% \begin{enumerate}
%     \item Each agent is assigned to exactly one task:
%     \[
%     \sum_{j=1}^{n} X_{ij} = 1, \quad \forall i \in \{1, \dots, n\}
%     \]

%     \item Each task is assigned to exactly one agent:
%     \[
%     \sum_{i=1}^{n} X_{ij} = 1, \quad \forall j \in \{1, \dots, n\}
%     \]

%     \item The assignment variables are binary:
%     \[
%     X_{ij} \in \{0,1\}, \quad \forall i,j \in \{1, \dots, n\}
%     \]
% \end{enumerate}

\begin{figure*}[htbp]
% \begin{figure*}[h!]
% \begin{figure*}[t!]
    \centering
    \begin{subfigure}{0.3\textwidth}
        \includegraphics[width=\linewidth]{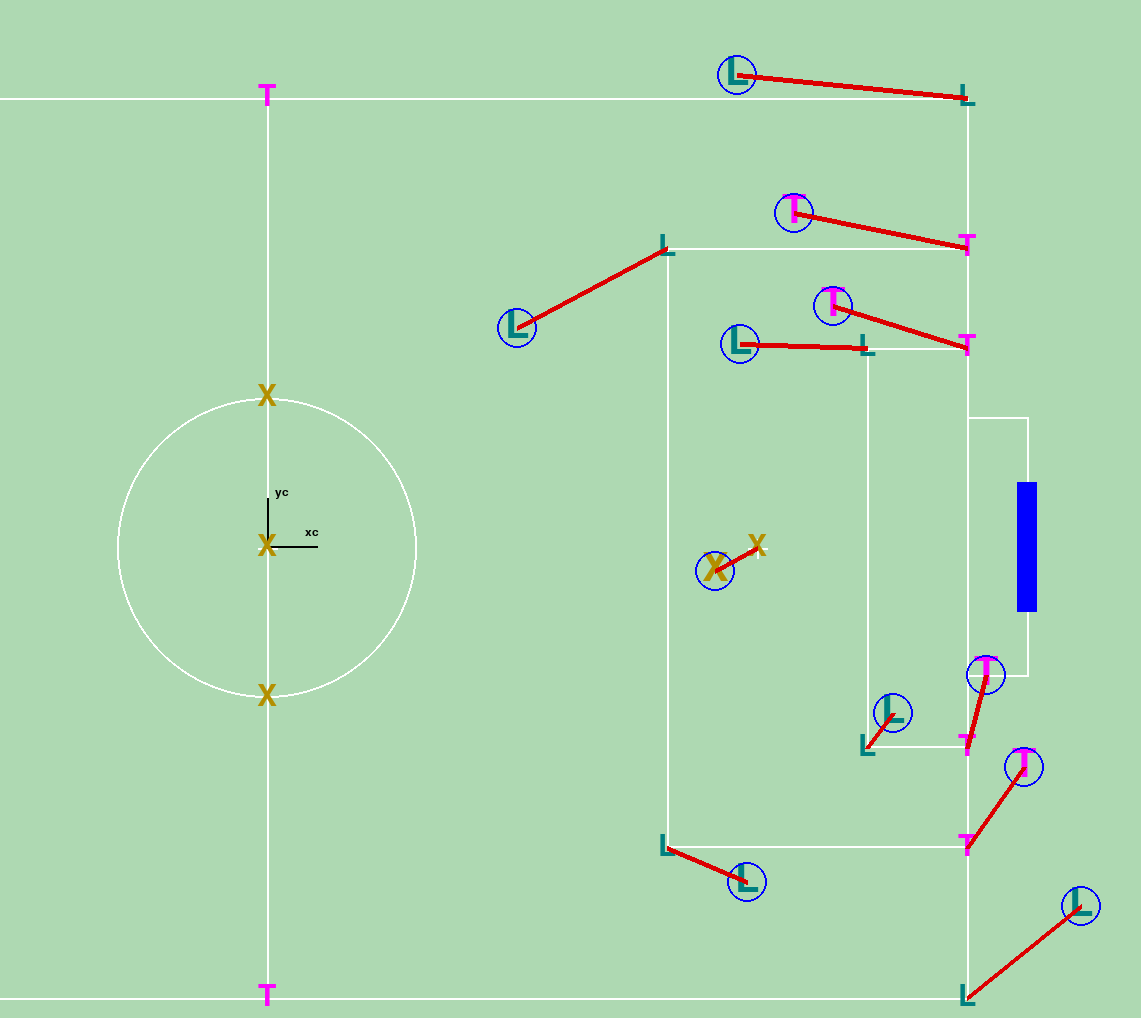}
        \caption{Matching landmarks separately with no misclassification.}
    \end{subfigure}
    \hfill
    \begin{subfigure}{0.3\textwidth}
        \includegraphics[width=\linewidth]{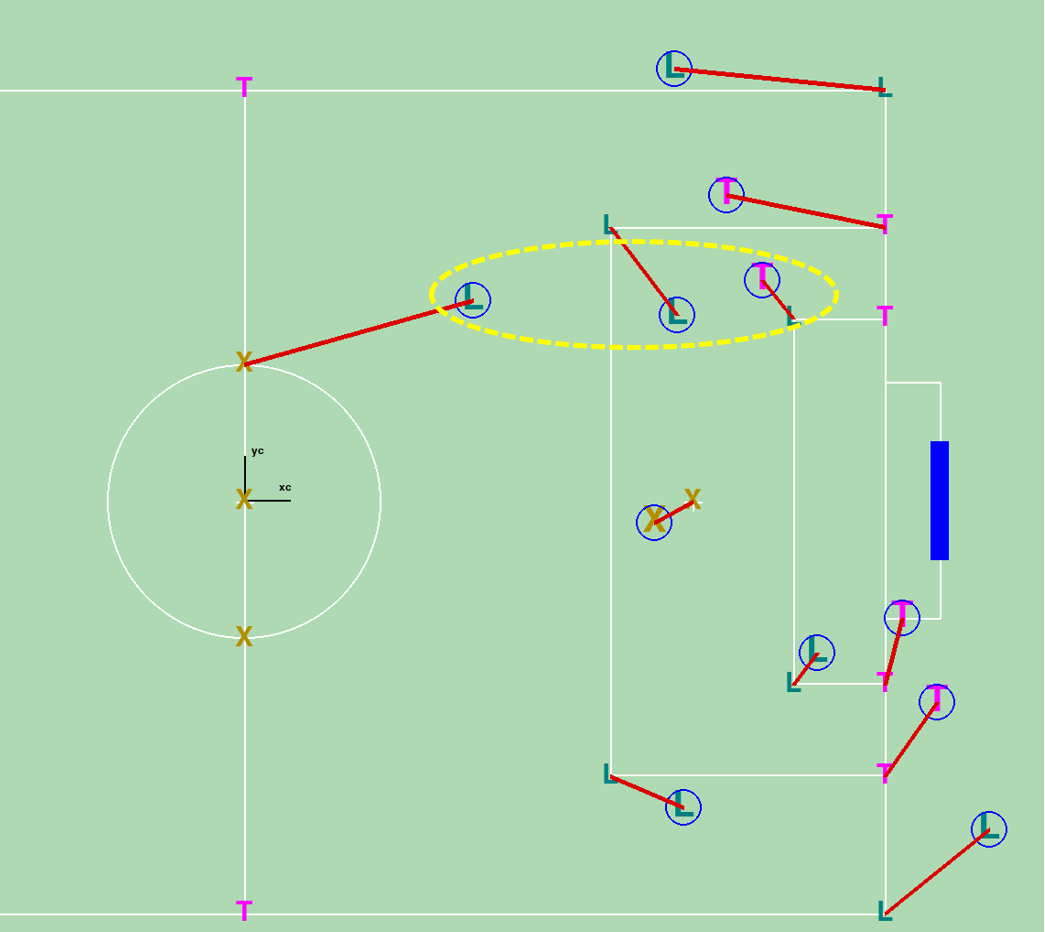}
        \caption{Matching landmarks identically with no misclassification.}
    \end{subfigure}
    \hfill
    \begin{subfigure}{0.3\textwidth}
        \includegraphics[width=\linewidth]{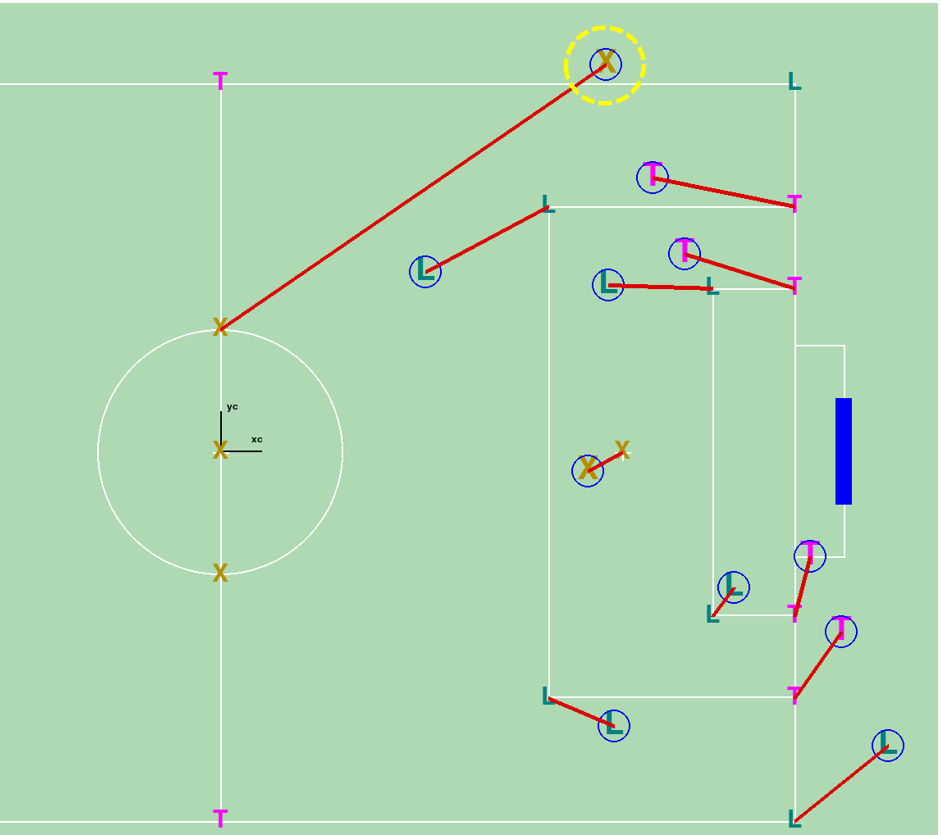}
        \caption{Matching landmarks separately with misclassification.}
    \end{subfigure}
    
    \caption{Landmark matching under different scenarios. This mismatched landmarks are circled in yellow. (a) shows the correct matching. (b) shows the mismatching due to treating different kinds of landmarks identically (c) shows the mismatching due to misclassification}
    \label{fig:Landmark_matching}
\end{figure*}

Different types of landmarks can be matched separately by creating distinct mappings based on their classification, or they can be treated identically by using the same mapping while ignoring their classification, as shown in \cref{fig:Landmark_matching}. Matching landmarks of different types separately generally improves accuracy when the classification is correct. However, if some landmarks are misclassified—commonly occurring with landmarks that are far away or in poor lighting conditions—the matching accuracy may decrease. Therefore, we match the different types of landmarks both separately and identically in parallel, and then choose the result with the least error.

% \FloatBarrier

The common methods for solving LAP are Kuhn–Munkres algorithm (Hungarian algorithm) \cite{kuhn1955hungarian,munkres1957algorithms}, Jonker-Volgenant algorithm \cite{jonker1988shortest}, and integer linear programming \cite{selmair2021evaluation}. We also note multiple improvements and modifications of the Jonker-Volgenant algorithm \cite{malkoff1997evaluation,crouse2016implementing}. 
% In the case that the number of data points is large, like when using LiDAR, nearest neighbour (NN) is also good for lower computational complexity \cite{bentley1975multidimensional,friedman1977algorithm,arya1998optimal}. 
% There are also papers comparing the performance of methods
The performance of methods is also compared in these papers
\cite{selmair2021evaluation,levedahl2000performance}.
% how to solve::

In our case, the number of points is relatively small, so what matters is guaranteeing the optimal solution in a fast enough time for real-time localization applications. Therefore, Kuhn–Munkres algorithm \cite{munkres1957algorithms}, Jonker-Volgenant algorithm \cite{jonker1988shortest} and the modified Jonker-Volgenant algorithm \cite{crouse2016implementing} are good candidates to use to solve LAP for our localization method. 

We simulated the soccer field and timed the three methods by randomly sampling 100,000 poses uniformly across the field. The results show that the modified Jonker-Volgenant algorithm \cite{crouse2016implementing} is the fastest, taking 0.0537 ms in average. Therefore, we chose this method to match our landmarks.

% \begin{figure}[htbp]
%     \centering
%     \includegraphics[width=\linewidth]{Figures/S2_data_assignment/matching_time_10000.png}
%     \caption{The solve time for three methods solving LAP.}
%     \label{fig:matching_time_10000}
% \end{figure}

\subsection{Pose Estimation}
\label{sec:Pose_Estimation}
% problem definition:
% \subsection{problem definition}
The next component of our localization is to estimate the 2D pose according to the matched landmarks from \cref{sec:Data_Association}. The 2D pose of a robot includes the position transformation \((t_x,t_y)\) and the orientation $\theta$. 

The problem is formulated as following.
Given a set of points in the body frame $p^{\text{body}}_{\text{i}}$ and the corresponding points in the world frame $p^{\text{world}}_{\text{i}}$, we need to find the 2D transformation \((t_x,t_y,\theta)\) which maps the body frame points to the world frame points.
We define the translation vector as $t = (t_{x},t_{y})$, and the rotation angle as $\theta$. The transformation can be expressed as:
\begin{equation*}
p^{\text{world}}_{\text{i}} = R(\theta) \cdot p^{\text{body}}_{\text{i}}+ t
\label{eq:2D_transformation}
\end{equation*}
where $R(\theta)$ is a 2D rotation matrix:
\begin{equation*}
R(\theta) = \begin{bmatrix} 
\cos(\theta) & -\sin(\theta) \\ 
\sin(\theta) & \cos(\theta) 
\end{bmatrix}
\label{eq:rotation_matrix}
\end{equation*}

% \[
% R(\theta) = \begin{pmatrix} \cos\theta & -\sin\theta \\ \sin\theta & \cos\theta \end{pmatrix}
% \]

Several methods that solve this include nonlinear programming (NLP), Direct Linear Transformation (DLT), and the Kabsch algorithm \cite{kabsch1976solution,arun1987least,maurer1996registration}. The NLP method is flexible, handling various types of transformations and constraints. However, the methods for NLP, such as Sequential Quadratic Programming, Interior-Point Method, and Levenberg-Marquardt method, are computationally expensive. Moreover, NLP might give a local optimal solution which can be detrimental for our use case. The DLT method and Kabsch algorithm provide closed-form solutions \cite{kabsch1976solution,arun1987least,maurer1996registration} for the 2D pose, which are faster and better meet our needs.

\section{Localization Framework}
\label{sec:localization_framework}

\subsection{Iterative Landmark Matching}

% \hl{Shouldn't this go right after the data assignment section? -> it needs pose_estimation}
% This is the core component of our localization algorithm. 
To correctly localize the robot, we need to match the landmarks detected correctly. The mismatching is a problem only if the initial guess of the 2D pose is far from the ground truth. One common method to deal with mismatching is MCL, which explores and evaluates particles in different poses, but it is computationally heavy. Our method for dealing with potential incorrect assignments is to iteratively correct the matching and remove potential outliers. 
% Our method to deal with the potential incorrect assignments is to correct the matching by iteratively removing potential outliers. 
% like Iterative Closest Point(ICP)\cite{besl1992method}. 

The algorithm is outlined in \cref{alg:iterative_landmark_matching}.
Given the initial guess for a pose $(t_{x_{0}}, t_{y_{0}}, \theta_0)$ and the landmark observation in the body frame $p^{\text{body}}$, calculate the corresponding position in world frame with an initial guess $p_{\text{guess}}^{\text{world}}$. Match the $p_{\text{guess}}^{\text{world}}$ to the map and find the landmark position in world frame $p^{\text{world}}$. Estimate the new pose $t_{x_i}, t_{y_i}, \theta_i$ as described in \cref{sec:Pose_Estimation}. Repeat the landmark matching and pose estimation process until it reaches a predetermined maximum iteration limit or until the new pose converges.

% Repeat the landmark matching and pose estimation process until reach the maximum iteration or the new pose is the same as old pose.

% algorithm here ::
\begin{algorithm}
\caption{Iterative Landmark Matching}
\label{alg:iterative_landmark_matching}
\begin{algorithmic}[1]
\State \textbf{Input:} $p^{\text{body}}$, $(t_{x_{0}}, t_{y_{0}}, \theta_0)$
\State \textbf{Output:} $(t_{x_{f}}, t_{y_{f}}, \theta_f)$
\State Initialize $p_{\text{guess}}^{\text{world}} = R(\theta_0)p^{\text{body}} + t_0$, counter = 1
\While{counter < max\_iteration}
    \State $p^{\text{world}}$ = \textbf{landmark\_matching}$(p_{\text{guess}}^{\text{world}})$ (\Cref{sec:Data_Association})
    \State $t_{x_i},t_{y_i},\theta_i$ =  \textbf{pose\_estimation}$(p^{\text{world}},p^{\text{body}})$ (\Cref{sec:Pose_Estimation})
    \If{$(t_{x_i}, t_{y_i}, \theta_i) = (t_{x_{i-1}}, t_{y_{i-1}}, \theta_i)$}
        \State break
    \EndIf
    \State counter = counter + 1
\EndWhile
\State \textbf{Return:} $(t_{x_{f}}, t_{y_{f}}, \theta_f)$
\end{algorithmic}
\end{algorithm}

We calculate the average calculation time for ILM with a maximum allowable iterations of 4. The average solving time for ILM using DLT is 0.901 ms and using Kabsh algorithm is 1.132 ms. Without considering the vision system, the ILM can be solved at approximately 1 kHz.

% \begin{table}[h!]
% \centering
% \caption{Average calculation time in millisecond for ILM with maximum allowable iterations of 4.}
% \label{tab:ILM_iter4_time}
% \begin{tabular}{|c|c|}
% \hline
% ILM using DLT & ILM using Kabsch algorithm \\
% \hline
% 0.901 ms & 1.132 ms  \\
% \hline
% \end{tabular}
% \end{table}

\subsection{Global Localization}
\begin{figure}[t]
    \centering
    \includegraphics[width= 0.9\linewidth]{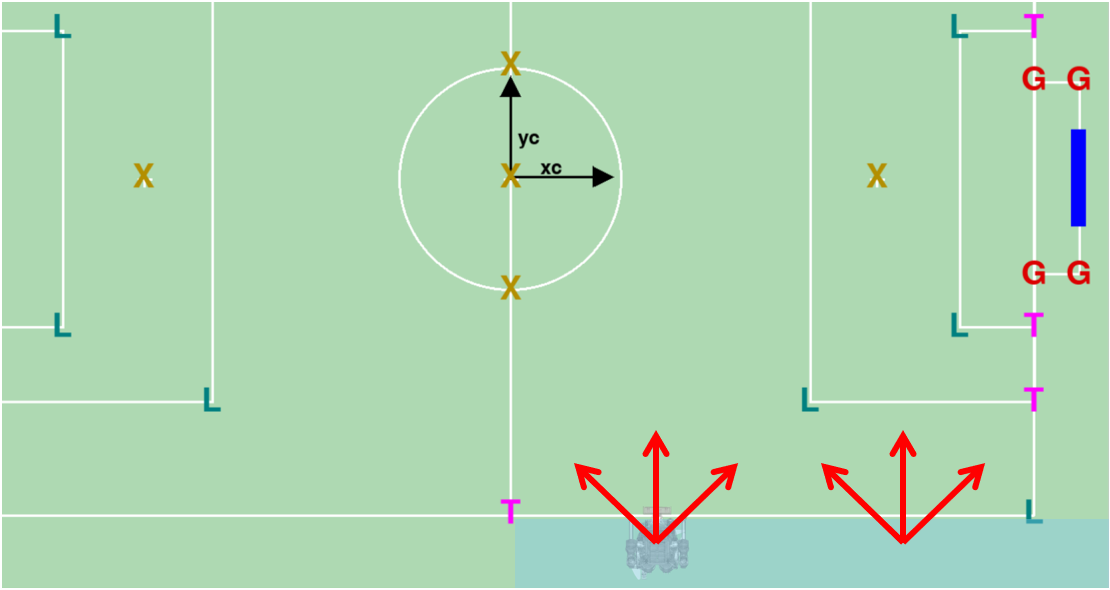}
    \caption{The multi-hypothesis representation of the starting pose. The blue shaded area indicates the potential robot starting region, while the red arrows illustrate the six hypothesized poses.}
    \label{fig:multi_hypothesis_localization}
\end{figure}

The soccer field is symmetric and the robot is not allowed to use magnetometers to find the direction. Therefore, robot will receive the same information in the symmetric pose.
According to the rules of RoboCup, the robot should start at the edge of our half side and face the field. Therefore, we use multi-hypothesis localization at the beginning of the match to find the pose. The initial hypothesis are shown in \cref{fig:multi_hypothesis_localization}. We use the frame which receives more than 5 landmarks and set a threshold of 0.5 meter to the maximum matching error. We calculate the initial pose using all the hypothesis and the pose with smallest matching error will then be chosen as the initial pose. The others are abandoned. As we will show in the experiment part, the ILM method has a high tolerance on the inital position and orientaion. Therefore, we do not need too much hypothesis. 

\subsection{Outlier Dropping}

\begin{figure}[htbp]
    \centering
    \includegraphics[width=0.9\linewidth]{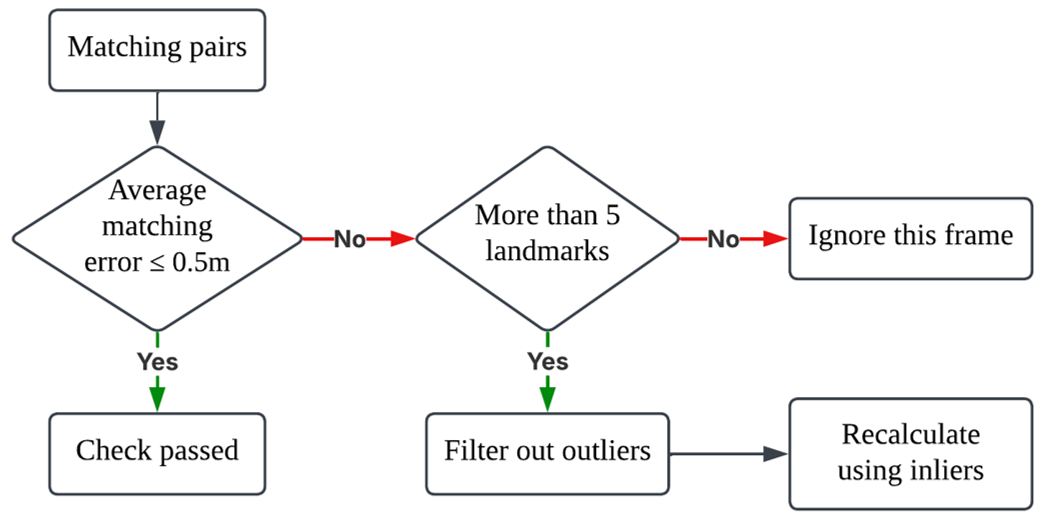}
    \caption{The outlier dropping process}
    \label{fig:outlier_dropping_process}
\end{figure}

Since the pose estimation will be influenced by wrong matching, we need to drop the outlier. \cref{fig:outlier_dropping_process} shows our outlier dropping process. It uses the average matching error to check if it needs outlier dropping. If the average matching error is larger than 0.5 m, we check if there are enough landmarks in the frame. If there are more than 5 landmarks, we use RANSAC \cite{fischler1981random} method to find the outlier and recalculate the pose using inliers.

% If the outlier rate is lower than 20\%, we use the thresholding to find outlier, which is faster. Otherwise, we check if there are more than 5 landmarks. If it has less landmark, then we ignore this frame. If it has more than 5 landmarks, we use RANSAC to find the outliers, which is slower but more robust than thresholding. The subset size is for RANSAC is 3 landmarks. 

\subsection{Filtering}
In the competition, rules allowed for only a single IMU and a camera. We use filtering to fuse IMU data with localized estimates. The velocity of the robot is estimated using  invariant extended Kalman filter (InEKF) \cite{ahn2023} on ARTEMIS. Two commonly used filter are Extended Kalman Filter (EKF) \cite{kim2008accurate} and Particle Filter (PF) \cite{chen2010particle,tariq20242d}. 

% When our robot gets new data from the IMU, it processes a prediction step. When ILM calculates the 2D pose from new vision data, it should process an update step \hl{What does this mean?}. 

Define the state vector \(X\) of the robot in 2D space and control input \(U\) as:
% \begin{align*}
% \mathbf{X_k} & :=
% \begin{bmatrix}
% t_{x_k} \\
% t_{y_k} \\
% \theta_k
% \end{bmatrix}
% &
% \mathbf{U_k} & :=
% \begin{bmatrix}
% v_{xk} \\
% v_{yk} \\
% \omega_{k}
% \end{bmatrix}
% \end{align*}
\begin{align*}
\mathbf{X_k} & := \begin{bmatrix} t_{x_k} & t_{y_k} & \theta_k \end{bmatrix}^T, \quad
\mathbf{U_k} := \begin{bmatrix} v_{fk} & v_{sk} & \omega_k \end{bmatrix}^T
\end{align*}

% Define the state vector $\mathbf{x}$ of the robot in 2D space is given by:
% \[
% \mathbf{x} = \begin{bmatrix}
% x \\
% y \\
% \theta
% \end{bmatrix}
% \]
where $t_x$ and $t_y$ represent the position of the robot, and $\theta$ represents its orientation.
% The control input vector $\mathbf{u}$ is:
% \[
% \mathbf{u} = \begin{bmatrix}
% v \\
% \omega
% \end{bmatrix}
% \]
$v_f$ is the forward velocity of the robot, $v_s$ is the side velocity and $\omega$ is the angular velocity.
% The continuous-time dynamics of the robot can be described by the following nonlinear equations:
% % \[
% % \dot{\mathbf{x}} = \begin{bmatrix}
% % \dot{x} \\
% % \dot{y} \\
% % \dot{\theta}
% % \end{bmatrix}
% % = \begin{bmatrix}
% % v \cos(\theta) \\
% % v \sin(\theta) \\
% % \omega
% % \end{bmatrix}
% % \]
% \begin{equation*}
% \begin{bmatrix}
% \dot{x} \\
% \dot{y} \\
% \dot{\theta}
% \end{bmatrix}
% =
% \begin{bmatrix}
% \cos(\theta) & -\sin(\theta) & 0 \\
% \sin(\theta) & \cos(\theta) & 0 \\
%  0 & 0 & 1
% \end{bmatrix}
% \begin{bmatrix}
% v_x \\
% v_y \\
% \omega
% \end{bmatrix}
% \end{equation*}
% where $\dot{x}$, $\dot{y}$, and $\dot{\theta}$ represent the time derivatives of $x$, $y$, and $\theta$, respectively.

For discrete-time implementation, assuming a time step $\Delta t$, the state update equations become:
% \[
% \mathbf{x}_{k+1} = \mathbf{x}_k + \begin{bmatrix}
% v_k \cos(\theta_k) \Delta t \\
% v_k \sin(\theta_k) \Delta t \\
% \omega_k \Delta t
% \end{bmatrix}
% + \mathbf{w}_k
% \]
\begin{equation*}
\mathbf{X}_{k+1} = \mathbf{X}_k 
+
\begin{bmatrix}
\cos(\theta_{k}) & -\sin(\theta_{k}) & 0 \\
\sin(\theta_{k}) & \cos(\theta_{k}) & 0 \\
 0 & 0 & 1
\end{bmatrix}
\mathbf{U}_k 
\Delta t
+
W_k
\end{equation*}
where $\mathbf{W}_k$ represents the process noise.
The measurement model is $\mathbf{Z}_{k} = \mathbf{X}_k + \mathbf{V}_k $, where \(V_k\) is measurement noise, since ILM gives the $\mathbf{X_k}$ directly.
Since the measurement model is linear and the dynamic model is weakly nonlinear, both EKF \cite{kim2008accurate} and PF \cite{chen2010particle,tariq20242d} work. We implmented particle filter in the RoboCup 2024 competition.

\section{experiment result} %%%%%%%%%%%%%%%%%%%%%%%%%%%%%%%%%%%%%%%%%%%%%%%%%%%%%%%%%%%%%%%%%%%%%%%%%%%%%%%%%%%%%%%%%%
\label{sec:exp_result}

\subsection{ILM vs ICP}
We compare the ILM with the ICP in simulation and find that ILM method is more robust towards the error in the initial guess and easier to get a correct matching.

\begin{figure*}[ht]
    \centering
    \includegraphics[width=0.8\linewidth]{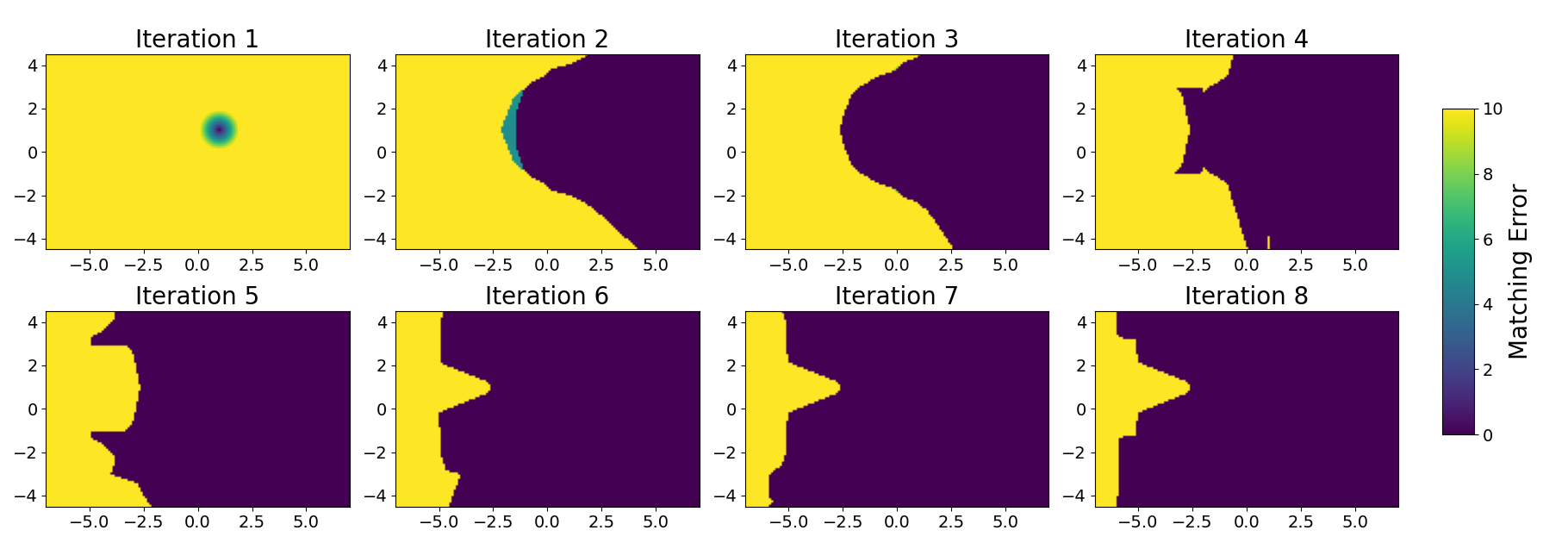}
    \caption{ Matching error of ILM using different initial guesses with different number of maximum allowable iterations. Dark purple means correct matching and bright yellow means incorrect matching}
    \label{fig:ilm_max_iter}
\end{figure*}

\begin{figure*}[ht]
    \centering
    \includegraphics[width=0.8\linewidth]{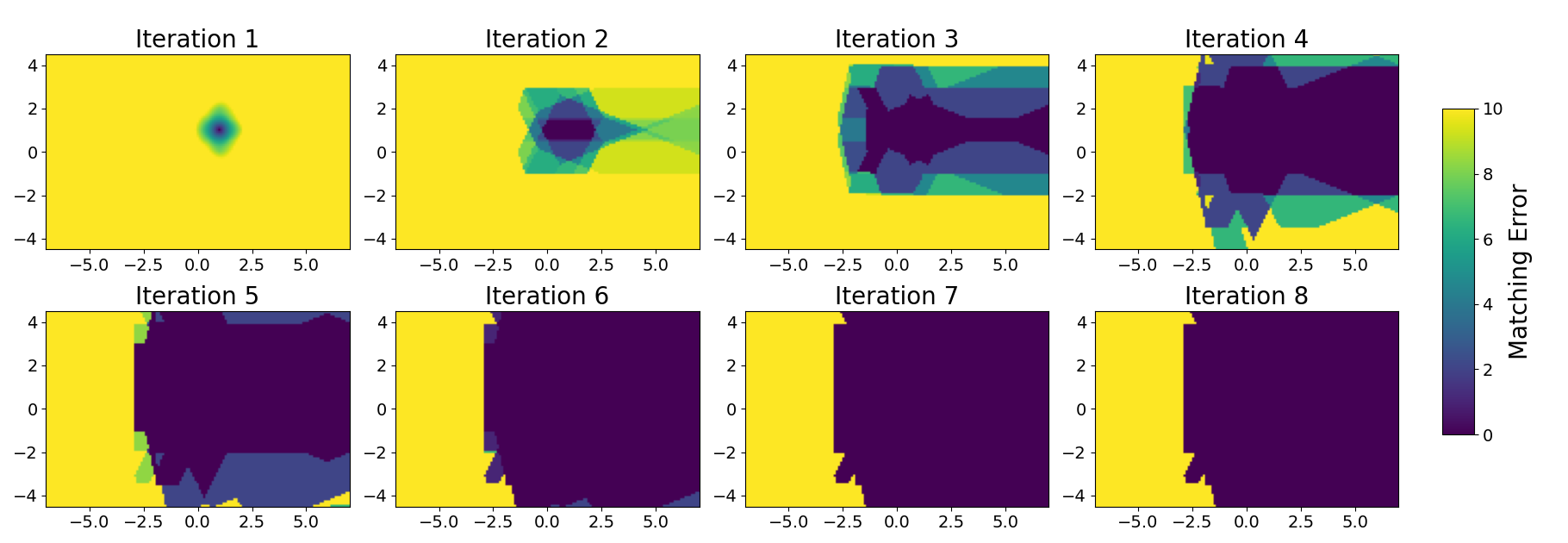}
    \caption{ Matching error of ICP using different initial guesses with different number of maximum allowable iterations. Dark purple means correct matching and bright yellow means incorrect matching}
    \label{fig:icp_max_iter}
\end{figure*}

\cref{fig:ilm_max_iter} shows the matching error under different initial guesses and different maximum allowable iterations for ILM. \cref{fig:icp_max_iter} shows the matching error using ICP. The true pose is \((t_x = 1,t_y = 1,\theta = 0)\). The initial guess \(t_{x_{0}}, t_{y_{0}}\) is equally sampled on the soccer field with \(\theta_{0}=0\). The heatmaps show the matching error where dark purple means correct matching and bright yellow means incorrect matching. 

\cref{fig:ilm_max_iter} demonstrates that after a small number of iterations (5 times), the region for correct matching dramatically expands and has a high probability for correct matching. Comparing iteration 1 and iteration 8, we see that without iterative matching, the data association is correct only when the initial guess is close to the ground truth \((t_x = 1,t_y = 1,\theta = 0)\). After iterating 8 times, the correct region covers most of the whole field (86.67\% of the field), which shows a large robustness for a wrong initial guess. The correct matching rate for random initial orientation for \((t_x = 1,t_y = 1,\theta = 0)\) is 52.78\% , which is also quite large. 

Comparing \cref{fig:ilm_max_iter} and \cref{fig:icp_max_iter}, the first 4 iterations show that ILM method finds the correct matching faster than ICP and the last 4 iterations show that the robustness for wrong initial guess using ILM is also larger than using ICP.  

% \begin{figure*}[ht]
%     \centering
%     \includegraphics[width=0.8\linewidth]{Figure_new/experiment/ILM_VS_ICP/heatmap_dist_err_data_array_2D_ilm.png}
%     \caption{Matching error of ILM using different initial guesses with different number of maximum allowable iterations. Dark purple means correct matching and bright yellow means incorrect matching}
%     \label{fig:ilm_max_iter}
% \end{figure*}

% \begin{figure*}[ht]
%     \centering
%     \includegraphics[width=0.8\linewidth]{Figure_new/experiment/ILM_VS_ICP/heatmap_dist_err_data_array_2D_icp.png}
%     \caption{Matching error of ICP using different initial guesses with different number of maximum allowable iterations. Dark purple means correct matching and bright yellow means incorrect matching}
%     \label{fig:icp_max_iter}
% \end{figure*}

\begin{figure}[t]
    \centering
    \includegraphics[width=1.0\linewidth]{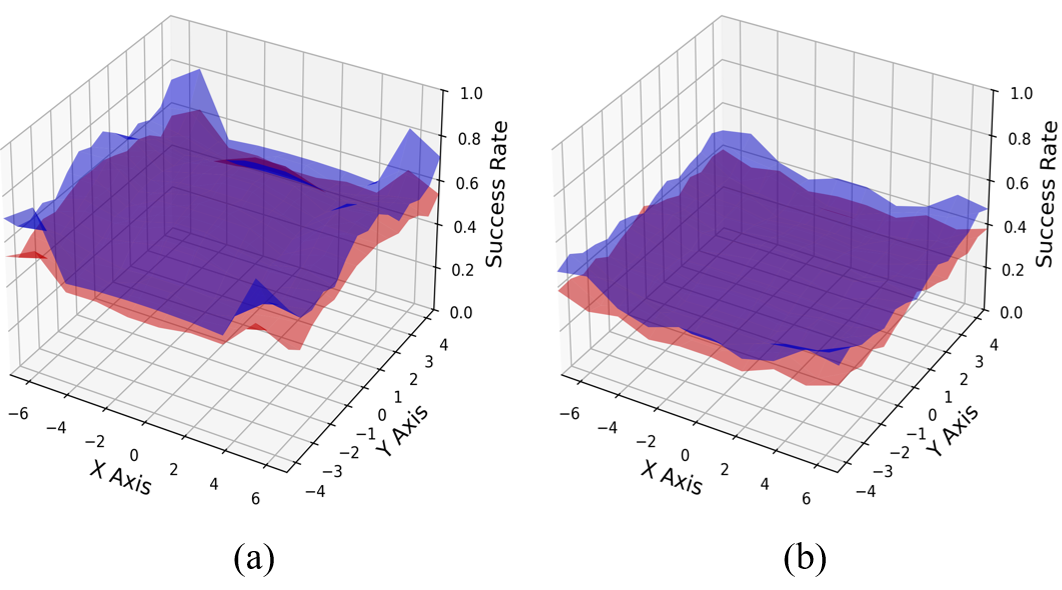}
    \caption{Correct matching rate with random initial guess. (a) Initial position guess; (b) Initial orientation guess. Blue surface for ILM and red surface for ICP.}
    \label{fig:matching_rate_random_compare}
\end{figure}

We do this analysis not only for the case \((t_x = 1,t_y = 1,\theta = 0)\) but also across the entire soccer field, comparing the ILM and ICP methods, where the true position and orientation are equally sampled. The correct matching rates with random initial positions and orientations are shown in \cref{fig:matching_rate_random_compare}. In both subfigures, the blue surface represents ILM, and the red surface represents ICP. In all cases, the blue surface is above the red surface, indicating that ILM demonstrates higher tolerance to errors in both initial position and orientation. 

% \begin{figure}[t]
%     \centering
%     \includegraphics[width=0.8\linewidth]{Figure_new/experiment/ILM_VS_ICP/fov_success_rate_compare_ilm.png}
%     \caption{Correct matching rate with random initial position guess. Blue surface for ILM and red surface for ICP. The correct matching rate of ILM is above ICP, showing higher tolerance on initial position guess.}
%     \label{fig:matching_rate_random_position}
% \end{figure}

% \begin{figure}[t]
%     \centering
%     \includegraphics[width=0.8\linewidth]{Figure_new/experiment/ILM_VS_ICP/fov_success_rate_compare_ilm_angle.png}
%     \caption{Correct matching rate with random initial orientation guess. Blue surface for ILM and red surface for ICP. The correct matching rate of ILM is above ICP, showing higher tolerance on initial orientation guess.}
%     \label{fig:matching_rate_random_orientation}
% \end{figure}

% \begin{figure}[t]
%     \centering
%     \begin{subfigure}[b]{0.75\linewidth}
%         \centering
%         \includegraphics[width=\linewidth]{Figure_new/experiment/ILM_VS_ICP/fov_success_rate_compare_ilm.png}
%         \caption{Correct matching rate with random initial position guess.}
%         \label{fig:matching_rate_random_position}
%     \end{subfigure}
    
%     \vspace{0.5cm} % Adjust vertical space between subfigures as needed
    
%     \begin{subfigure}[b]{0.75\linewidth}
%         \centering
%         \includegraphics[width=\linewidth]{Figure_new/experiment/ILM_VS_ICP/fov_success_rate_compare_ilm_angle.png}
%         \caption{Correct matching rate with random initial orientation guess.}
%         \label{fig:matching_rate_random_orientation}
%     \end{subfigure}
    
%     \caption{Comparison of correct matching rates with random initial position and orientation guesses for ILM and ICP. Blue surface for ILM and red surface for ICP.  The correct matching rate of ILM is above ICP, showing higher tolerance on initial position and orientation guess.}
%     \label{fig:matching_rate_comparison}
% \end{figure}

\subsection{Simulation}
We simulate a rectangular trajectory where the vertices are the corners of goal region. Uniformly sampled noise of \(\pm\)0.5 m is added to landmark observations. Uniformly sampled noise of \(\pm\)0.02 m is added to the position and \(\pm\)0.02 rad for orientation every 10 ms. 

\begin{figure}[t]
    \centering
    \includegraphics[width=0.9\linewidth]{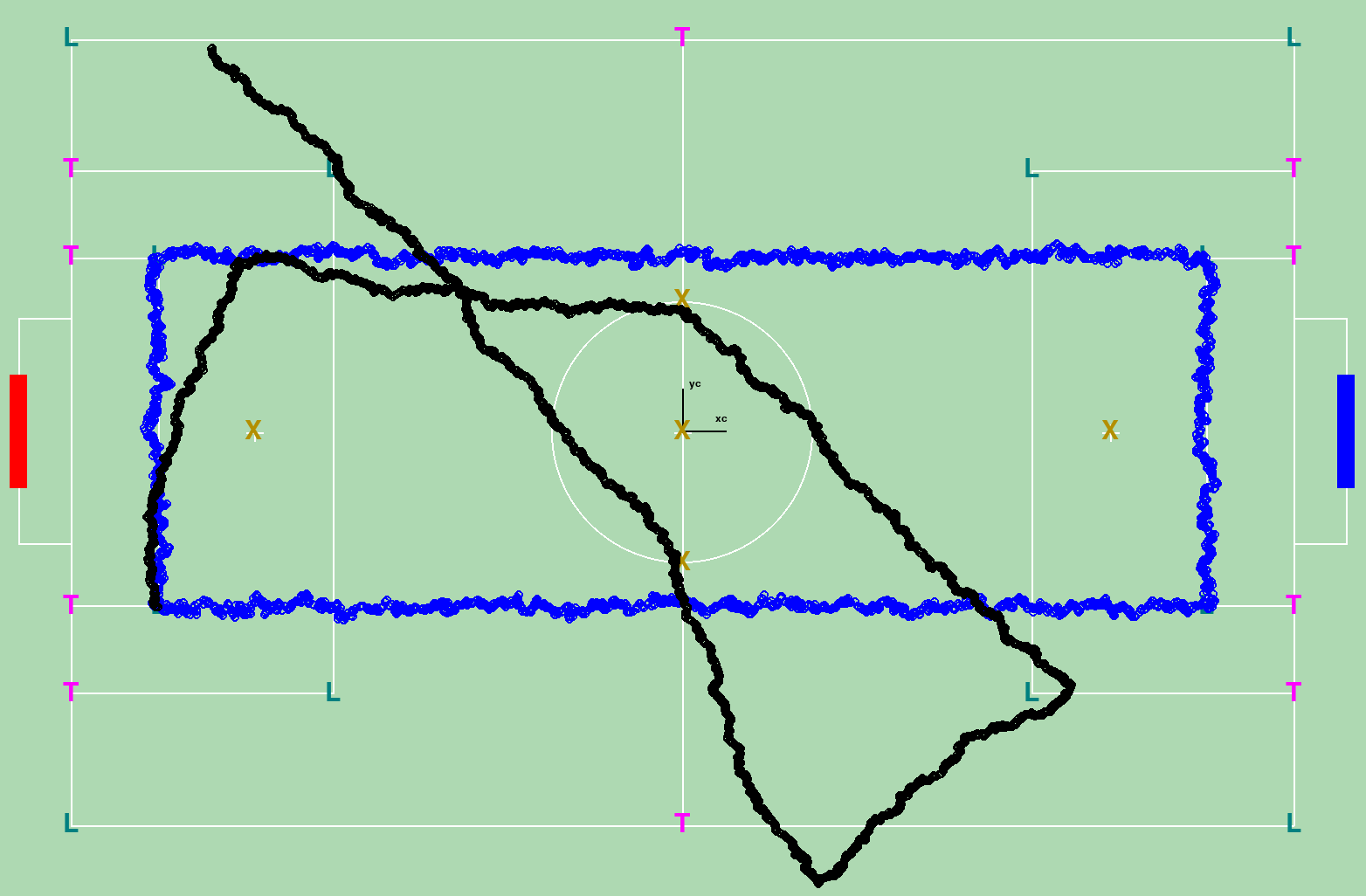}
    \caption{Simulated trajectory using ILM (blue line) and only IMU (black line) for state estimation of a trajectory touching all 4 corners of the goal box.}
    \label{fig:sim_ILM_vs_imu}
\end{figure}

\begin{figure}[t]
% \begin{figure*}[h!]
% \begin{figure*}[t!]
    \centering
    \begin{subfigure}{0.225\textwidth}
        \includegraphics[width=0.8\linewidth]{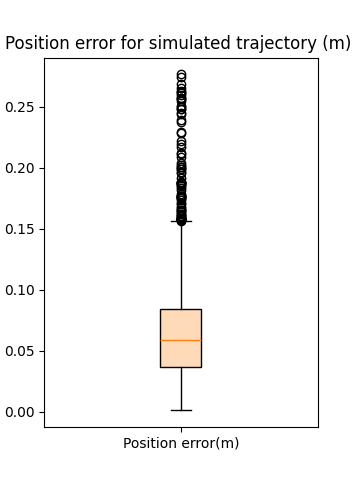}
        % \caption{matching landmarks separately,no misclassification}
    \end{subfigure}
    \hfill
    \begin{subfigure}{0.254\textwidth}
        \includegraphics[width=0.8\linewidth]{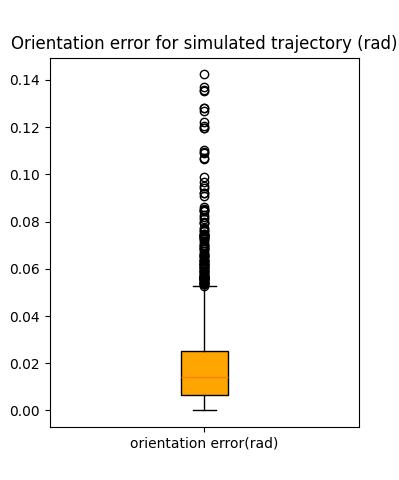}
        % \caption{matching landmarks identically,no misclassification}
    \end{subfigure}    
    \caption{Position and orientation errors for simulation trajectory.}
    \label{fig:sim_statics}
\end{figure}

\cref{fig:sim_ILM_vs_imu} depicts a trajectory with our method in blue and a trajectory purely with IMU data alone in black. The position and orientation error is shown in \cref{fig:sim_statics}. We can see that our localization method works well under large sensor noise. 

\subsection{Field Test}
% In \cref{fig:real_ILM_vs_imu}, we manually controlled ARTEMIS \cite{ahn2023} shown in \cref{fig:artemis_shooting} to follow a trajectory as closely as possible to the one shown in \cref{fig:sim_ILM_vs_imu}. The robot completed three laps at an approximate speed of 0.3 m/s. Despite real-world disturbances, the actual trajectory remained relatively close to the simulated trajectory in \cref{fig:sim_ILM_vs_imu}. However, the estimate relying solely on the IMU continued to drift over time.

\begin{figure}[b]
    \centering
    \includegraphics[width=1.0\linewidth]{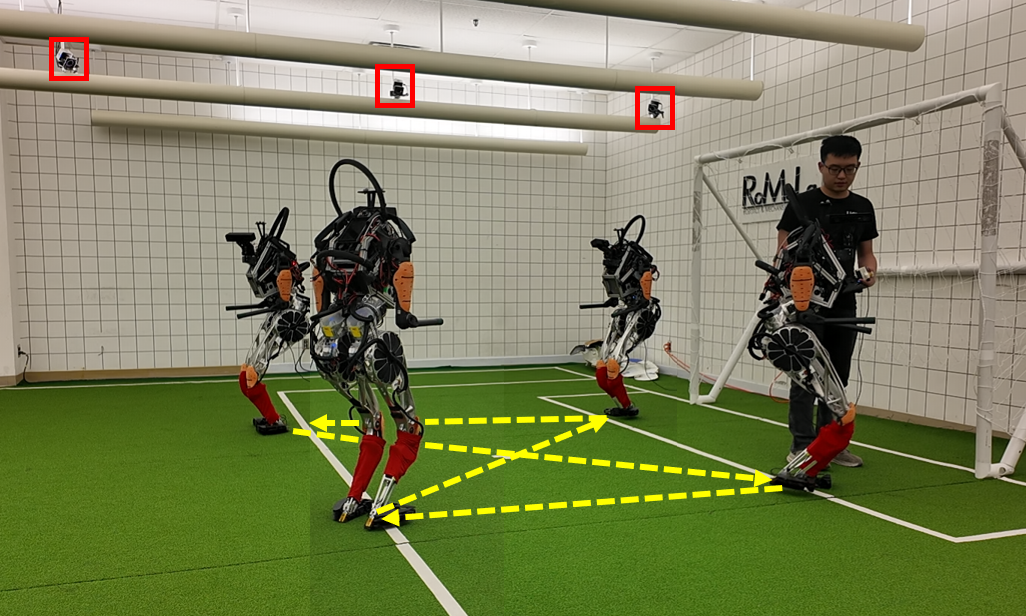}

    \caption{A combined figure from different frames shows that ARTEMIS autonomously followed the X-shaped trajectory, indicated by yellow dashed lines. Motion capture cameras, enclosed in red boxes, provided the ground truth for the robot's pose.}
    \label{fig:field_test}
\end{figure}

\begin{figure}[htbp]
    \centering
    \begin{subfigure}[b]{0.15\textwidth}
        \includegraphics[width=\textwidth]{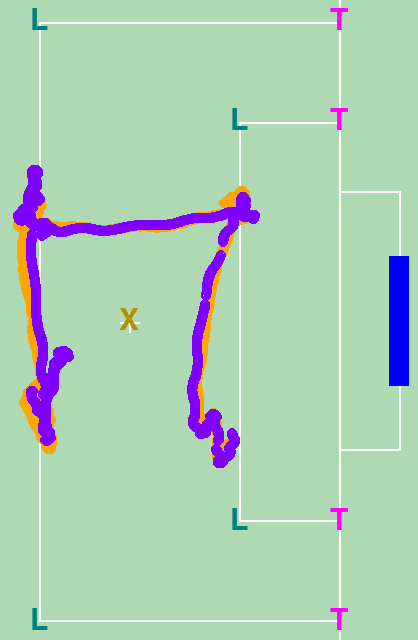}
        \caption{}
        \label{fig:a}
    \end{subfigure}
    \hfill
    \begin{subfigure}[b]{0.15\textwidth}
        \includegraphics[width=\textwidth]{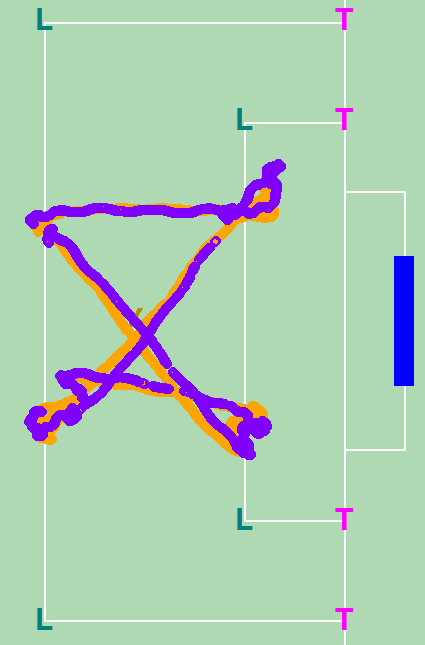}
        \caption{}
        \label{fig:b}
    \end{subfigure}
    \hfill
    \begin{subfigure}[b]{0.157\textwidth}
        \includegraphics[width=\textwidth]{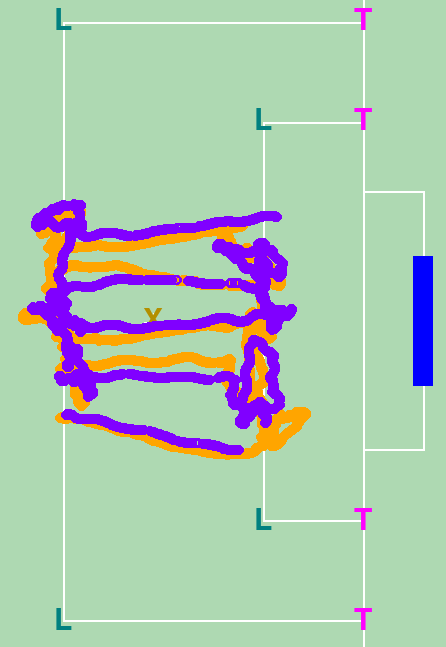}
        \caption{}
        \label{fig:c}
    \end{subfigure}
    \caption{Three tested trajectories in a real soccer field. (a): C-shape pattern; (b): X-shape pattern; (c): Zigzag pattern. The localization result is shown in orange, while the ground truth position from motion capture is in bright purple.}
    \label{fig:three_trajectories}
\end{figure}

In \cref{fig:field_test}, ARTEMIS autonomously followed the desired trajectories using our path planning and following algorithm \cite{hou2025path}. Three different types of trajectories were tested to verify the accuracy and robustness of our proposed localization method, as shown in \cref{fig:three_trajectories}. The trajectories appear unsmooth due to trajectory following errors and the periodic oscillations of the robot body in the bipedal locomotion. The robot completed five laps of each trajectory at a maximum speed of 1.0 m/s, with only the first lap shown for clarity. The ground truth pose was provided by the Vicon motion capture system in the lab. Due to the limited coverage area of the mocap system, the three trajectories do not cover the entire soccer field. Localization results are shown in orange, while the ground truth position is shown in bright purple. Some parts of the red line are disconnected due to mocap losing track of the markers.   

% \begin{figure}[htbp]
%     \centering
%     \begin{subfigure}{0.32\textwidth}
%         \centering
%         \includegraphics[width=\linewidth]{Figure_new/experiment/field_test/trajectory_1.png}
%         \caption{Trajectory 1}
%     \end{subfigure}
%     \vfill
%     \begin{subfigure}{0.32\textwidth}
%         \centering
%         \includegraphics[width=\linewidth]{Figure_new/experiment/field_test/trajectory_2.png}
%         \caption{Trajectory 2}
%     \end{subfigure}
%     \vfill
%     \begin{subfigure}{0.32\textwidth}
%         \centering
%         \includegraphics[width=\linewidth]{Figure_new/experiment/field_test/trajectory_3.png}
%         \caption{Trajectory 3}
%     \end{subfigure}
%     \caption{Field test trajectories. Red is ground truth position from mocap, Blue is position from ILM.}
%     \label{fig:three_trajectories}
% \end{figure}

% \begin{figure}[t]
%     \centering
%     \includegraphics[width=0.9\linewidth]{Figure_new/experiment/field_test/traj_total.png}

%     \caption{Three tested trajectories in a real soccer field. (a): C-shape pattern; (b): X-shape pattern; (c): Zigzag pattern. The localization result is shown in blue, while the ground truth position from motion capture is in red.}
%     \label{fig:three_trajectories}
% \end{figure}

\cref{tab:error_statistics} presents the Root Mean Square Error (RMSE), along with the minimum and maximum errors in position and orientation across the three trajectories. For comparative analysis, we implemented Augmented Monte Carlo Localization (aMCL) as introduced by Kim and Min \cite{kim2023enhancing}, noted for its accuracy in real-time localization within the RoboCup environment. The particle count was configured to 200. Our localization method demonstrates accuracy comparable to that of aMCL, with slight improvements observed in certain instances. However, aMCL's average computation time is 0.0114 seconds (approximately 87.7 Hz), rendering it approximately ten times slower than our approach. Additionally, in scenarios where MCL particles exhibit significant divergence across the field, transitioning to the flipped pose is feasible due to the symmetric nature of the soccer field.

\begin{table}[t]
    \centering
    \caption{Error statistics for different trajectories comparing ILM and aMCL.}
    \label{tab:error_statistics}
    
    \begin{subtable}[t]{\linewidth}
        \centering
        \caption{Position Error Statistics}
        \label{tab:position_error}
        \begin{tabular}{l|cc|cc|cc}  
            \toprule
            \multirow{2}{*}{Trajectory} & \multicolumn{2}{c|}{\makecell{Position \\ RMSE \\ (meter)}} & 
            \multicolumn{2}{c|}{\makecell{Minimum \\ Position Error \\ (meter)}} & 
            \multicolumn{2}{c}{\makecell{Maximum \\ Position Error \\ (meter)}} \\
            \cmidrule(lr){2-3} \cmidrule(lr){4-5} \cmidrule(lr){6-7}
            & ILM & aMCL & ILM & aMCL & ILM & aMCL \\
            \midrule
            Trajectory 1 & 0.214  & 0.241  & 0.002  & 0.001  & 0.558  & 0.713  \\
            Trajectory 2 & 0.191  & 0.294  & 0.002  & 0.003  & 0.437  & 0.639  \\
            Trajectory 3 & 0.185  & 0.253  & 0.003  & 0.007  & 0.503  & 0.598  \\
            \bottomrule
        \end{tabular}
    \end{subtable}
    
    \vspace{1em} % Adds vertical space between subtables

    \begin{subtable}[t]{\linewidth}
        \centering
        \caption{Orientation Error Statistics}
        \label{tab:orientation_error}
        \begin{tabular}{l|cc|cc|cc}  
            \toprule
            \multirow{2}{*}{Trajectory} & \multicolumn{2}{c|}{\makecell{Orientation \\ RMSE \\ (degree)}} & 
            \multicolumn{2}{c|}{\makecell{Minimum \\ Orientation Error \\ (degree)}} & 
            \multicolumn{2}{c}{\makecell{Maximum \\ Orientation Error \\ (degree)}} \\
            \cmidrule(lr){2-3} \cmidrule(lr){4-5} \cmidrule(lr){6-7}
            & ILM & aMCL & ILM & aMCL & ILM & aMCL \\
            \midrule
            Trajectory 1 & 3.189  & 3.282  & 0.0013  & 0.0007  & 8.316  & 12.127  \\
            Trajectory 2 & 4.092  & 4.197  & 0.0007  & 0.0013  & 12.974  & 14.673  \\
            Trajectory 3 & 3.231  & 3.012  & 0.0003  & 0.0004  & 12.218  & 16.050  \\
            \bottomrule
        \end{tabular}
    \end{subtable}

\end{table}

\section{Conclusion} %%%%%%%%%%%%%%%%%%%%%%%%%%%%%%%%%%%%%%%%%%%%%%%%%%%%%%%%%%%%%%%%%%%%%%%%%%%%%%%%%%%%%%%%%%%%%
\label{sec:conclusion}
%%% What did we learn
% describe theory
In this paper, we presented a fast, accurate, and robust vision-based 2D localization method capable of operating at approximately 1 kHz, achieving a positional RMSE of 0.2 m and an orientation RMSE of 3.5 degrees. This approach enables ARTEMIS to accurately estimate its real-time pose within a predefined soccer field. Field experiments demonstrated that our method outperformed adaptive Monte Carlo Localization (aMCL) with 200 particles in both accuracy and computational efficiency. Furthermore, comparisons between ILM and Iterative Closest Point (ICP) revealed that ILM exhibited greater robustness to errors in the initial pose estimate, leading to a higher rate of correct landmark matches.

\end{document}